\documentclass{article}

\usepackage{arxiv}

\usepackage[utf8]{inputenc} % allow utf-8 input
\usepackage[T1]{fontenc}    % use 8-bit T1 fonts
\usepackage{hyperref}       % hyperlinks
\usepackage{url}            % simple URL typesetting
\usepackage{booktabs}       % professional-quality tables
\usepackage{amsfonts}       % blackboard math symbols
\usepackage{nicefrac}       % compact symbols for 1/2, etc.
\usepackage{microtype}      % microtypography
\usepackage{lipsum}		% Can be removed after putting your text content
\usepackage{graphicx}
\usepackage[round]{natbib}
\usepackage{doi}
\usepackage{threeparttable}
\usepackage{bm}

\title{Enhance Information Propagation for Graph Neural Network by Heterogeneous Aggregations}

\author{
  Dawei ~Leng \thanks{The first and corresponding author}, \hspace{2mm}Jinjiang Guo,  \hspace{2mm}Lurong Pan,  \hspace{2mm}Jie Li, \hspace{2mm} Xinyu Wang \\
 \emph{AIDD Group}\\
 Global Health Drug Discovery Institute, Beijing, China \\
  \texttt{dawei.leng@ghddi.org, jinjiang.guo@ghddi.org, lurong.pan@ghddi.org, jie.li@ghddi.org} \\

}

\date{}

\hypersetup{
pdftitle={Enhance Information Propagation for Graph Neural Network by Heterogeneous Aggregations},
pdfauthor={Dawei Leng},
pdfkeywords={GNN, HAG-Net},
}

\begin{document}
\maketitle

\begin{abstract}
	Graph neural networks are emerging as continuation of deep learning success w.r.t. graph data. Tens of different graph neural network variants have been proposed, most following a neighborhood aggregation scheme, where the node features are updated via aggregating features of its neighboring nodes from layer to layer. Though related research surges, the power of GNNs are still not on-par-with their counterpart CNNs in computer vision and RNNs in natural language processing. We rethink this problem from the perspective of information propagation, and propose to enhance information propagation among GNN layers by combining heterogeneous aggregations. We argue that as richer information are propagated from shallow to deep layers, the discriminative capability of features formulated by GNN can benefit from it. As our first attempt in this direction, a new generic GNN layer formulation and upon this a new GNN variant referred as HAG-Net is proposed. We empirically validate the effectiveness of HAG-Net on a number of graph classification benchmarks, and elaborate all the design options and criterions along with.
\end{abstract}

\section{Introduction}

Success of deep learning in computer vision and natural language processing has recently boosted flood of research on applying neural networks to graph data \citep{survey}. Graph is a simple yet versatile data structure jointly described by sets of nodes and edges. Aside from image and text data we're familiar, lots of real world data are better described as graph and thus processed by graph neural networks, such as social networks \citep{social}, financial fraud detection \citep{fraud}, knowledge graph \citep{KG}, biology interaction network \citep{PPI}, small molecule in drug discovery \citep{gine}, to name a few. 

Since the seminal works \citep{gcn,graphsage}, tens of different graph neural network variants have been proposed, emphasizing different graph properties and design options. GNN research routes can be roughly divided into two categories: spectral based and spatial based. Spectral based GNNs try to approximate CNN's convolution by defining Fourier transform on graph \citep{gcn} and thus where the name graph convolution network comes from. The major limitation of spectral based GNNs is that graph convolution is defined on a fixed global graph, thus not suitable for tasks where graph structure is changing from sample to sample. Meanwhile, most recent works take the spacial-based direction \citep{gin,gine,gunet,gat,NNconv}, i.e., GNNs process the graph data following a neighborhood aggregation scheme, where the node features at layer $k$ are updated via aggregating features of its neighboring nodes from layer $k-1$.

GNN's neighborhood aggregation is a direct imitation of CNN's convolution in spatial dimensions. However, unlike image in computer vision where data reside on regular grid, graph data are intrinsically orderless. For an undirected graph $G = (V, E)$ there're only \textit{nodes} $V$ and \textit{edges} $E$ defined. By "orderless", it means from the perspective of any center node $v\in V$, there's no way to tell which node $u\in \mathcal{N} (v)$ is its $n$-th neighbor, where $\mathcal{N}(v)$ stands for the neighborhood of node $v$. This special property of graph impedes GNNs from replicating CNN's success on image data and RNN's success on sequential data. 

Most work in spatial-based GNNs can be formally summarized by the following two-stage architecture \citep{survey, gine}: first propagate node information among each other by neighborhood aggregation, then form the whole graph representation by a read-out function. The $k$-th layer of a spatial-base graph neural network is like
\begin{equation} \label{eq:aggr} h_v^{(k)}=\mathbf{C}^{(k)}(h_v^{(k-1)}, \mathbf{A}^{(k)}(\{(h_v^{(k-1)}, h_u^{(k-1)}, e_{uv}), u\in \mathcal{N}(v)\})) \end{equation}
in which $h_v^{(k)}$ is the feature vector of node $v$ at the $k$-th layer, $e_{uv}$ is the edge feature vector between node $u$ and $v$, and $\mathcal{N}(v)$ is usually the one-hop neighbors. Function $\mathbf{A(\cdot)}$ aggregates node features in the neighborhood of $v$, and $\mathbf{C(\cdot)}$ combines features both from center node and aggregated from its neighborhood by $\mathbf{A(\cdot)}$. To obtain the entire graph’s representation $h_G$, the READOUT function pools node features from the final iteration $K$ as
\begin{equation} \label{eq:readout} h_G=\mathbf{R}(\{h_v^{(K)}| v \in V\}) \end{equation}

Normally a spatial-base graph neural network consists of a stack of multiple aggregation layers and finally one readout layer. Published works differ in either $\mathbf{A(\cdot)}$, $\mathbf{C(\cdot)}$ or $\mathbf{R(\cdot)}$, among which aggregation function $\mathbf{A(\cdot)}$ is the most important part because it determines how information propagates among nodes. Due to the orderless property of graph data, function $\mathbf{A(\cdot)}$ must be permutation-free, i.e., for any given center node, all its neighbor nodes must be treated equally. This drastically restricts possible choices for the aggregation function, commonly $\{max, min, mean, sum, mul, att\}$ in which $att$ is an attention operator as in \citep{gat}. The permutation-free restriction is also true for read-out function $\mathbf{R(\cdot)}$.

Usually one certain neighborhood aggregation operator is chosen for GNN design. For example, the seminal work GraphSAGE \citep{graphsage} takes the layer formulation as \begin{equation} \label{eq:sage} h_v^{(k)}=\phi_1(h_v^{(k-1)}) + \phi_2(mean({h_u^{(k-1)}, u\in\mathcal{N}(v)})) \end{equation} where $\mathbf{A(\cdot)}=mean({h_u^{(k-1)}, u\in\mathcal{N}(v)})$, $\mathbf{C(\cdot)}=sum(\phi_1, \phi_2)$ and neighborhood aggregation takes the $mean$ operator. GIN \citep{gin} argues that $mean$ operator loses neighborhood size information, and proposes the layer formulation as \begin{equation} \label{eq:gin} h_v^{(k)}=\phi((1+\epsilon)h_v^{(k-1)}+\sum_{u\in\mathcal{N}(v)}h_u^{(k-1)}) \end{equation} where $\mathbf{A(\cdot)}=\sum_{u\in\mathcal{N}(v)}h_u^{(k-1)})$, $\mathbf{C(\cdot)}=\phi(sum(1+\epsilon, 1)$ and neighborhood aggregation takes the $sum$ operator. As further improvement, \citep{gine} incorporates edge features $e_{u,v}$ by taking the layer formulation as \begin{equation} \label{eq:gine} h_v^{(k)}=\phi((1+\epsilon)h_v^{(k-1)}+\sum_{u\in\mathcal{N}(v)}ReLU(h_u^{(k-1)}+e_{u,v}) \end{equation} where $\mathbf{A(\cdot)}=\sum_{u\in\mathcal{N}(v)}ReLU(h_u^{(k-1)}+e_{u,v})$, $\mathbf{C(\cdot)}=\phi(sum(1+\epsilon, 1)$ and neighborhood aggregation also takes the $sum$ operator. \citep{NNconv} takes the layer formulation as \begin{equation} \label{eq:nnconv} h_v^{(k)}=\phi_1(h_v^{(k-1)})+aggr_{u\in\mathcal{N}(v)}h_u^{(k-1)}\phi_2(e_{u,v}) \end{equation} in which neighborhood aggregation $aggr$ takes either $mean$, $max$ or $sum$ operator.

From the perspective of information propagation, when only permutation-free operators are allowed, information loss after neighborhood aggregation is always inevitable, since there's no way to differentiate among neighbor nodes, and consequently, it won't be possible to recover the input graph $G$ by aggregation result $G'$. Consider the commonly used $mean$ operator, after each aggregation layer, features from different nodes are averaged, the result graph $G'$ will be a blurred version of the input graph $G$. Deeper the graph neural network, more blurred the result is. With such a lossy intermediate representation, it'll be hard for the neural network to fulfill the downstream learning task effectively. 

In this manuscript we try to improve GNN's performance from the perspective of enhancing information propagation from shallow to deep layers. We argue that as richer information are propagated, the discriminative capability of features formulated by GNN can benefit from there. As our first attempt in this direction, we propose to enhance information propagation by combining heterogeneous aggregations in function $\mathbf{A}(\cdot)$. The underlining philosophy is straightforward: each aggregation operator extracts/describes different aspect of the input graph $G$, by combining different aggregation operators, the information propagation loss can be mitigated, thus allowing more effective features for downstream task to propagate to deep layers. With this in mind, a new generic GNN layer formulation and upon this a new GNN variant referred as HAG-Net is proposed. We empirically validate the effectiveness of HAG-Net on a number of graph classification benchmarks, and elaborate all the design options and criterions along with. We focus on graph-level tasks here whereas the same technique can be also applied to node-level tasks without any difficulty.

\begin{figure}
	\centering
	\includegraphics[width=1.0\textwidth]{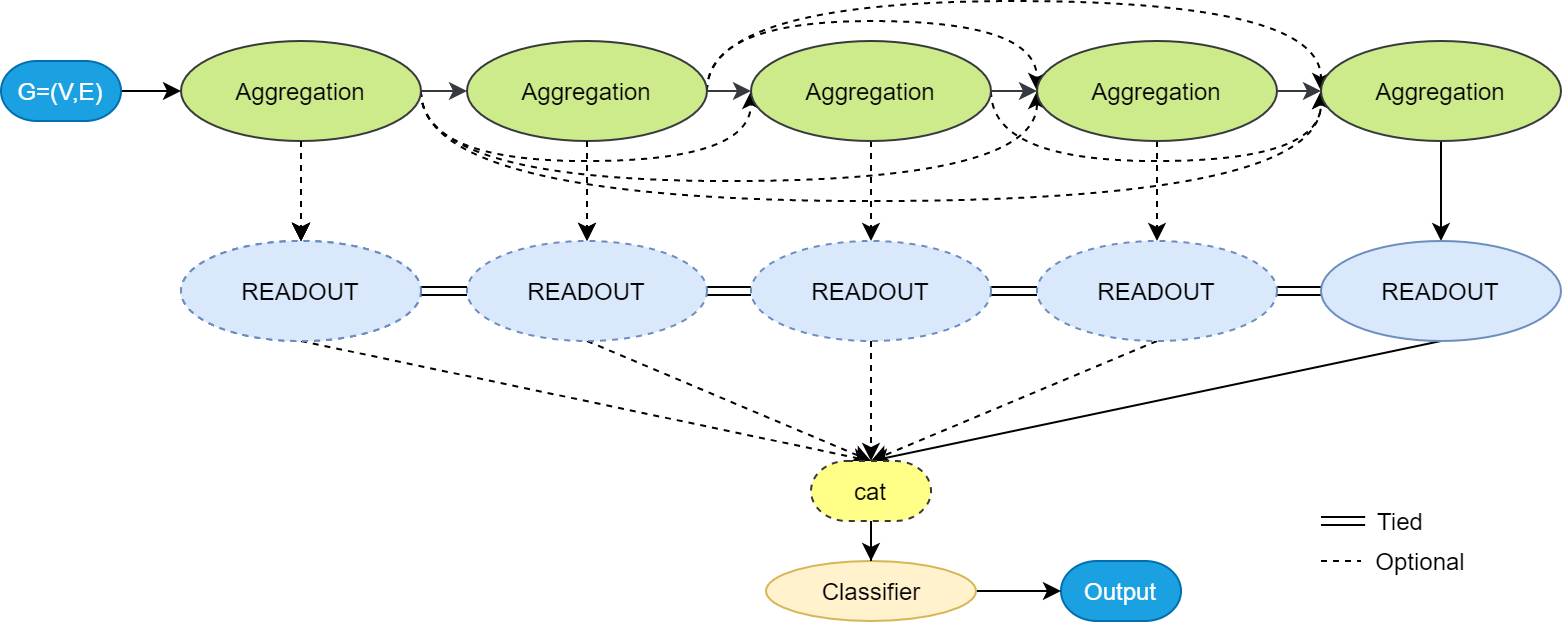}
	\caption{Network structure of HAG-Net. The proposed network is designed to be generic, the pyramid feature stacking by multiple READOUT layers and dense connection among neighborhood aggregation layers are optional. The neighborhood aggregation layer is formulated by eq.\ref{eq:haggr} and READOUT layer is formulated by eq.\ref{eq:hreadout}}
	\label{fig:HAGNet}
\end{figure}

\section{Method}

We begin by reformulating eq.\ref{eq:aggr}. For graph $G=(V, E)$ with $v,u \in V$ and $e_{uv}\in E$, the generic layer formulation with heterogeneous aggregations is as \begin{equation} \label{eq:haggr} h_v=\psi(\mathbf{C}(h_v, \bigoplus_i^{M-1}\phi_i(\mathbf{A}_{i,u\in \mathcal{N}(v)}(\{(h_v, h_u, e_{uv})\})))) \end{equation} In which $\mathbf{A}_i(\cdot), i=0\cdots M-1$ are $M$ different aggregation operators,  $\bigoplus$ is the merge operator for $M$ neighborhood aggregation results, and $\mathbf{C}(\cdot)$ updates center node $v$'s feature with the merged aggregation result. $\psi$ and $\phi_i$ are linear/non-linear transform functions, and layer index $k$ is omitted here for conciseness, the left-hand computation is always about layer $k-1$ if not specified otherwise. 

For node-level tasks, the node representation $h_v$ from the final aggregation layer is usually used for prediction. For graph-level tasks, the READOUT function aggregates node features from the final aggregation layer to form the entire graph’s representation as \begin{equation} \label{eq:hreadout} h_G=\psi(\bigoplus_i^{M-1}\phi_i(\mathbf{A}_{i,v\in V}(\{h_v\}))) \end{equation} Compared to eq.\ref{eq:readout}, the READOUT function here also benefits from the enhanced information propagation by heterogeneous aggregations. Note for READOUT function the aggregations are performed over all the graph nodes instead of local neighborhood.

\subsection{Design Options}

\paragraph{Aggregation operators} Commonly used aggregation operators include $\{max, sum, mean, att\}$. Aforementioned $\{mul, min\}$ operators though satisfying the permutation-free restriction, could easily cause computation instability issues, thus seldom used for neighborhood aggregation in practice. $sum$ and $mean$ operators act similarly, though $sum$ is theoretically preferred over $mean$ \citep{gin} because it keeps the neighborhood size information, there's latent risk that computation overflow could happen for large graph, where $mean$ is a much safer choice. $mean(\mathcal{N}(v))$ is actually the 0-order statistical moment of $\mathcal{N}(v)$, higher order moments such as $variance$, $skewness$ and $kurtosis$ also meet the permutation-free restriction. $max$ operator is widely used in neural networks in fields such as computer vision and natural language processing, where it usually acts as pooling function. $att$ from \citep{gat} is a transformer like multi-head self-attention for $\mathcal{N}(v)$ following by a $sum$ operator, thus it acts more like $sum(\varphi(\mathcal{N}(v)))$. In this manuscript we focus on operator set $\{max, sum, mean, att\}$, and leave other operators for future investigation.

\paragraph{Merge heterogeneous aggregations} For $\bigoplus$ operator in eq.\ref{eq:haggr} and \ref{eq:hreadout}, there're two commonly used options $\{cat, sum\}$ in which $cat$ means concatenation. The transform function $\phi_i$ after each aggregation result is commonly implemented by a stack of dense layers.

\paragraph{Update center node features} In eq.\ref{eq:haggr}, center node features $h_v$ will be updated with information aggregated from neighborhood by function $\mathbf{C}(\cdot)$. Possible choices include $\{sum, max, cat, rnn\}$, in which $rnn$ \citep{gated} is an RNN cell for example \verb+LSTMCell+ or \verb+GRUCell+ in Pytorch with merged neighborhood aggregation result as hidden state and center node feature as input. Note this sequential setup is totally artificial, their roles for the RNN cell can be exchanged at will. We leave $mean$ method out here because it acts quite similar with $sum$, possibly due to the following transform $\psi$ and batch normalization built-in. Like $\phi_i$, The transform function $\psi$ is commonly implemented by a stack of dense layers.

\subsection{Model Structure}

With the generic neighborhood aggregation layer as eq.\ref{eq:haggr} and READOUT layer as eq.\ref{eq:hreadout}, we build our graph neural network HAG-Net for graph-level tasks by stacking multiple neighborhood aggregation and READOUT layers. The simplest structure is just a sequential stacking a multiple neighborhood aggregation layers and one READOUT layer, the output of the READOUT layer will be used as the representation for the whole graph. Former works such as \citep{gin,gunet} implement complex structures by using features from intermediate layers. We follow this idea, and design optional pyramid feature structure and dense connection among intermediate layers. The complete network structure is illustrated in Figure \ref{fig:HAGNet}.

For READOUT layers in Figure \ref{fig:HAGNet}, we empirically find that downstream task performance can benefit marginally from restricting their weights tied if pyramid structure is enabled. The downstream task classifier is implemented as a stack of 3 dense layers.

With heterogeneous aggregations the neighborhood aggregation layer and READOUT layer could be implemented as numerous variants with different options. Plus there're variation options within the model structure of HAG-Net itself. To determine these options, we treat them as model hyper-parameters and tune them by human expert as well as automatic grid search, the details are described in section \ref{sec: hyperpara}.

\section{Experiments}

\begin{figure}
	\centering
	\includegraphics[width=0.73\textwidth]{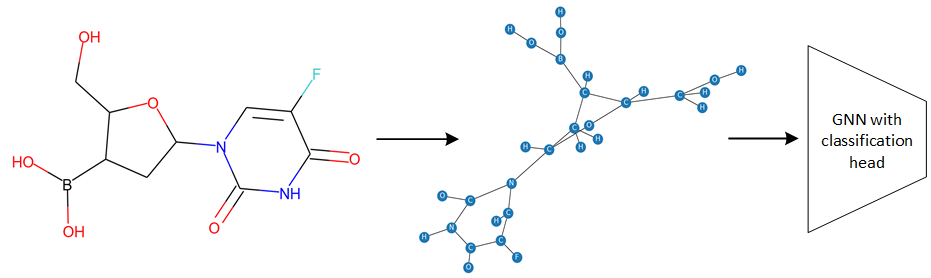}
	\caption{A molecule is converted to graph representation by treating each atom as node and covalent bond between atoms as edge, then a GNN with classification head is used to predict its activity. For general purpose different covalent bonds are all modeled into the same typed edges. With this we can do comparison among different GNN models, though it's not the most accurate way to model a molecule.}
	\label{fig:graphclassification}
\end{figure}

In this section we will try to answer the following questions through experiments:
\begin{itemize}
	\item Q1. Whether combining heterogeneous aggregations would improve GNN's performance? 
	\item Q2. What aggregation operator combination is the best, will larger combination always gear up model performance? 
	\item Q3. How to choose proper aggregation operator combination in practice? 
\end{itemize}
For real-world case study, in the following we'll conduct experiments on drug discovery datasets, where GNN is used to predict whether an organic small molecule is active w.r.t. certain biology target. A molecule is converted to graph representation by treating each atom as node and covalent bond between atoms as edge, see Figure \ref{fig:graphclassification}. We didn't choose benchmark datasets used previously in \citep{gin, gunet, graphsage, lanc} due to their limited sample size, usually less than 1K. All GNN models we've tested present large performance variation with different data splitting and weights initialization on them. Instead, we collect 5 datasets from drug discovery industry with sample size >5K. All these datasets are binary, vary both in size (from 7K to 76K) and class distribution (from balanced to highly biased). The details are given below.

\subsection{Dataset}

\paragraph{Antivirus7k} This is the smallest among all the 5 datasets, with 7,305 samples in total, and P/N ratio = 0.49 / 0.51. These're phenotypic records of antiviral bioactivity from various species and \textit{in vitro} assays, collected from commercial database. We choose EC50 <= 100nM as cutoff threshold.

\paragraph{CYP2C9V12k} This dataset is from recently published TDC project \citep{tdc}, with 12,092 samples in total and P/N ratio = 0.67 / 0.33. The CYP P450 genes are involved in the formation and breakdown (metabolism) of various molecules and chemicals within cells. Specifically, the CYP P450 2C9 plays a major role in the oxidation of both xenobiotic and endogenous compounds \citep{cyp2c9}.

\paragraph{Malaria16k} This dataset contains 16,933 samples in total with P/N ratio = 0.93 / 0.07. It's curated from public dataset of Malaria sensitivities assays \citep{malaria} by removing molecule chirality and merging consequent duplicates by \verb+OR+ operation.

\paragraph{Mtb18k} This is a public dataset with 18,886 samples in total, and P/N ratio = 0.88 / 0.12. These're phenotypic records of \textit{in vitro} assays against Mycobacterium tuberculosis \citep{mtb}. We choose activity cutoff of 10 $\mu$M.

\paragraph{Globalvirus76k} This is the biggest among all the 5 datasets, with 76,247 samples in total and P/N ratio = 0.57 / 0.43. These're combination of target based and phenotype based records of antiviral bioactivity from various species and \textit{in vitro} assays, collected from commercial database. We choose EC50 <= 100nM as cutoff threshold.

\subsection{Metrics} \label{sec:metrics}

With all the datasets being about binary classification task, commonly ER and AuROC metrics are used to evaluate model performance. Whereas for dataset with highly biased class distribution, AuROC will be dominated by the major class, thus we propose to use AuPR metric instead for model evaluation, specifically, the harmonic average of AuPRs of both positive and negative classes are reported in the following experiments.

There's a less studied topic in GNN research but we find important in production: convergence stability. During our evaluation of various GNN models, we notice that some models present much higher variance than others along the convergence progress, see Figure \ref{fig:convergencevar} for an illustration. Model with large convergence variance is not reliable in production environment since evaluation data is always limited. To measure the model convergence variance quantitatively, we propose to use median filtering method. For any convergence metric curve $x$, we first smooth the curve by median filtering and then compute the standard variance of their difference, i.e., \begin{equation} \label{eq:mstd} mstd = std(x - f_{median}^w(x)) \end{equation} in which $f_{median}^w$ is the median filter with window size = $2*w+1$. We set $w=5$ for all the experiments.

\subsection{Hyper-parater Determination} \label{sec: hyperpara}

The proposed HAG-Net is a generic structure supporting numerous variants with different combinations of design options. With this versatility, we explore answers to aforementioned questions Q1, Q2 and Q3. We tune hyper-parameters on \textbf{Antivirus7k} dataset because its suitable size and balanced class distribution, then results are reported on all the 5 datasets studied.

For answer to Q3 first, we tune hyper-parameters by human expert at the initial stage. We choose to copy GIN \citep{gin} structure for HAG-Net as baseline. Better hyper-parameter combinations are then explored by human expert heuristically. This process converges to configuration set \textit{cfg1} in Table \ref{tab:cfg} after dozens of trials. Evaluation results of HAG-Net with configuration set \textit{cfg1} on all the 5 datasets are given in Table \ref{tab:result}. 

We then further tune the hyper-parameters by automatic grid search. We use \textit{cfg1} as start point, and resort to Optuna \citep{optuna} package for the searching. This process runs for approximately one week on a cluster with 8 P100 GPUs. The hyper-parameter set with the best AuPR value is selected as the final decision, refer to configuration set \textit{cfg2} in Table \ref{tab:cfg} for details. And the whole process answers question Q3.

For questions Q1 \& Q2, by comparing \textit{cfg2} with \textit{cfg1} and evaluation results in Table \ref{tab:result}, we can see that GNN's performance can surely benefit from combining heterogeneous aggregations, but this benefit is not always consistent with the combination size. In the human expert optimized \textit{cfg1}, both $\{\mathbf{A}_{aggr}\}$ and $\{\mathbf{A}_{RO}\}$ are combinations of 2 different aggregation operators; whereas in grid search optimized \textit{cfg2}, for $\{\mathbf{A}_{RO}\}$, single operator $max$ is chosen over previous combination of $\{max, sum\}$. And during the hyper-parameter tuning process, we notice that for both $\{\mathbf{A}_{aggr}\}$ and $\{\mathbf{A}_{RO}\}$, combination with $\geq3$ aggregation operators never wins out. One possible reason might be that the investigated operator set $\{max, sum, mean, att\}$ here is quite limited, thus only marginal complementary effect could be obtained from larger combination. 

\begin{table}
	\centering
	\caption{Hyper-parameter configuration set for HAG-Net$^*$}
	\begin{threeparttable}
	\centering
	\begin{tabular}{llllllllll}
		\toprule
		Name     & \# ALayer  & $\{\mathbf{A}_{aggr}\}$  &  $\bigoplus_{aggr}$  & $\mathbf{C}$ &  $\{\mathbf{A}_{RO}\}$ &  $\bigoplus_{RO}$ & Pyramid & RO Tied & DC \\
		\midrule
		% cfg0     & 4  & $sum$ & - & $sum$ & $sum$ & - & True & False & False   \\
		cfg1     & 5  & $max, mean$  & $sum$ & $rnn$ & $max,sum$ & $sum$ & True & True & True   \\
		cfg2     & 5  & $max, sum$  & $sum$ & $cat$ & $max$ & - & True & True & False   \\
		\bottomrule
	\end{tabular}
	\label{tab:cfg}
	\begin{tablenotes}
        \footnotesize
		\item[*] "\# ALayer" is the number of aggregation layers, $\{\mathbf{A}_{aggr}\}$ is the set of aggregation operators in eq.\ref{eq:haggr} and $\{\mathbf{A}_{RO}\}$ is the set of aggregation operators in eq.\ref{eq:hreadout}.  $\bigoplus_{aggr}$ is the merge operator in eq.\ref{eq:haggr} and  $\bigoplus_{RO}$ is the merge operator in eq. \ref{eq:hreadout}. When "Pyramid" is True, there will be READOUT layer attached to each aggregation layer as in Figure \ref{fig:HAGNet}, othewise it will be a total sequential structure. "RO Tied" indicates whether the weights of READOUT layers are tied. "DC" indicates whether there are dense connections among aggregation layers.
	\end{tablenotes}
	\end{threeparttable}
\end{table}

\subsection{Performance Comparison}

In this subsection we benchmark the performance of the proposed HAG-Net with different hyper-parameter configurations as in Table \ref{tab:cfg}. Results from multiple state-of-the-art models such as GIN \citep{gin}, GUNet \citep{gunet}  and model from DeepChem \citep{deepchem} are also reported on the 5 datasets studied. GIN is a very popular model according to the leaderboard of OGB project \citep{ogb}. For GUNet we use $\{\mathbf{A}_{RO}\} = \{sum\}$ and the same classifier as in GIN. The model from DeepChem is specially designed for small moleclue tasks, both in network structure and input node features.

All the models are implemented with Pytorch 1.6.0. For model from DeepChem, node/atom features specially customized from DeepChem itself is used, which is of dimension $d = 75$. For GIN, GUNet and proposed HAG-Net, a node embedding layer is utilized to learn features from training data automatically, with also dimension $d = 75$ for comparison consistency. For GIN, GUNet and DeepChem models, default values are kept for their hyper-parameters, and Adam \citep{adam} optimizer with learning rate 1e-3 is used. For HAG-Net with different hyper-parameter configurations, SGD optimizer with learning rate 1e-2 is used. The training batch size and epoch number is fixed to 256 and 1,000 for all experiments. Evaluation results are given in Table \ref{tab:result}, reported as 5-fold average.

\begin{table}
	\centering
	\caption{Evaluation Results: 5-Fold Average$^*$}
	\begin{threeparttable}
	% \scalebox{0.8}{
	\resizebox{\textwidth}{!}{
	\begin{tabular}{@{}lllllllllll@{}}
	\toprule
	 Model        & \multicolumn{2}{l}{Antivirus7k}       & \multicolumn{2}{l}{CYP2C9V12k}         & \multicolumn{2}{l}{Mtb18k}                & \multicolumn{2}{l}{Malaria16k}       & \multicolumn{2}{l}{Globalvirus76k} \\ 
	 \midrule
	 GUNet        & $80.6\pm0.5$      & $24.8\pm1.0$      & $82.3\pm2.0$      & $19.9\pm1.0$       & $50.2\pm7.1$ &  $11.9\pm0.2$              & $\bm{69.0}\pm0.1$ & $6.9\pm0.1$      & $76.3\pm1.5$  & $27.7\pm0.6$           \\
	 GIN          & $91.1\pm0.6$      & $14.6\pm0.5$      & $84.6\pm0.9$      & $18.1\pm0.7$       & $71.1\pm1.8$ &  $10.0\pm0.2$              & $64.2\pm2.5$ & $6.3\pm0.2$           & $92.5\pm0.7$  & $13.3\pm0.4$           \\
	 DeepChem     & $91.2\pm0.8$      & $13.9\pm1.3$      & $85.7\pm0.7$      & $17.2\pm0.5$       & $73.6\pm1.9$ &  $\ \ 9.5\pm0.4$           & $67.3\pm1.4$ & $\bm{6.1}\pm0.1$      & $93.7\pm0.3$  & $12.8\pm0.1$           \\
	 \cmidrule(r){1-11}
	 HAG-Net cfg1 & $92.2\pm0.8$      & $12.7\pm0.4$      & $86.6\pm1.3$      & $\bm{16.4}\pm1.0$  & $75.9\pm0.7$ &  $\ \ 9.3\pm0.2$           & $66.1\pm3.4$ & $6.7\pm0.3$           & $94.9\pm0.2$  & $11.7\pm0.4$           \\
	 HAG-Net cfg2 & $\bm{92.5}\pm0.5$ & $\bm{12.0}\pm0.4$ & $\bm{87.1}\pm0.4$ & $\bm{16.4}\pm0.1$  & $\bm{77.1}\pm1.2$ & $\ \ \bm{9.0}\pm0.3$  & $67.8\pm3.3$ & $6.4\pm0.3$           & $\bm{95.6}\pm0.2$  & $\bm{10.8}\pm0.3$   \\
	 \bottomrule
	\end{tabular}}
	\label{tab:result}
	\begin{tablenotes}
        \footnotesize
		\item[*] For each dataset AuPR and ER results are reported, both in range $[0.0, 100.0]$.
	\end{tablenotes}
	\end{threeparttable}
\end{table}

\subsection{Discussion and Future Work}
\begin{figure}
	\centering
	\includegraphics[width=0.95\textwidth]{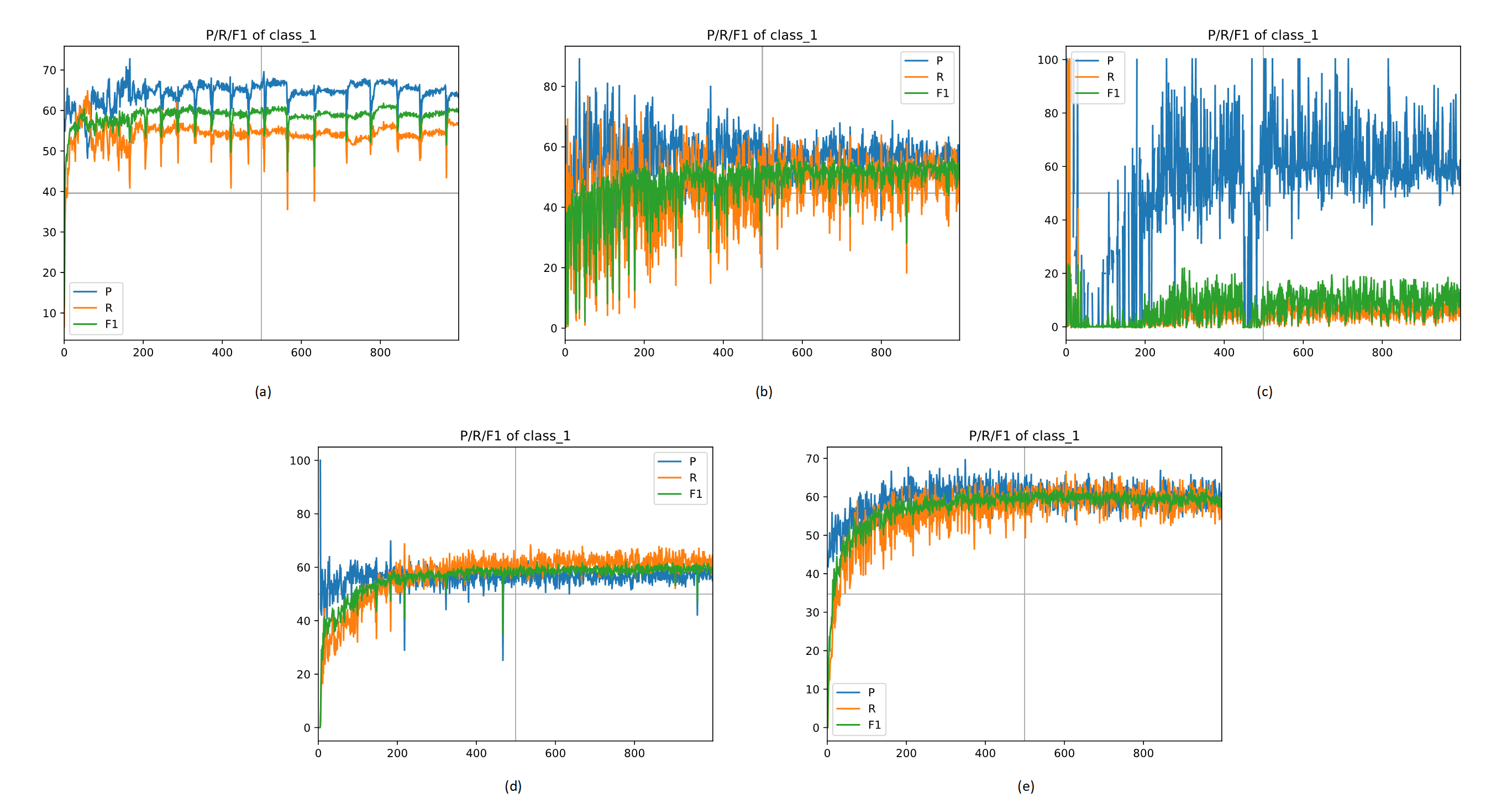}
	\caption{Convergence curves of P/R/F1 for class 1 on Mtb18k dataset. (a) DeepChem model (b) GIN model (c) GUNet model (d) HAG-Net cfg1 (e) HAG-Net cfg2}
	\label{fig:convergencevar}
\end{figure}

\paragraph{Convergence stability} As mentioned in section \ref{sec:metrics}, GNN models exhibit strikingly different stability during the convergence process. For GIN, GUNet and DeepChem models we studied, for different dataset and different weights initialization, frequent spikes and ditches can be observed in metric curves (with different patterns, see Figure \ref{fig:convergencevar} for illustration). This problem is not observable with the single point metrics such as ER, AuPR as reported in Table \ref{tab:result} since model with large convergence variance can still achieve high AuPR score. We use the $mstd$ metric defined in eq.\ref{eq:mstd} to measure the model convergence variance quantitatively, results are reported in Table \ref{tab:mstd} for ER curve, also as 5-fold average. Note the $mstd$ metric is still not perfect for evaluation of convergence stability, for example as illustrated by Figure \ref{fig:convergencevar}.(a), DeepChem model exhibits spurious fluctuation meanwhile achieves the smallest $mstd$ value.

\paragraph{Future work} Results reported in Table \ref{tab:result} and \ref{tab:mstd} validate our conjecture that GNN's performance can benefit from enhanced information propagation from shallow to deep layers. Our first attempt by combining heterogeneous aggregations succeeds, but not significantly. One possible future work is to investigate more neighborhood aggregation operators. 

Another possible method for enhancing information propagation is to use multi-channelling mechanism as in CNN. The channel dimension is essential for the success of CNN, where each channel encodes different part of information from the input samples. Combining heterogeneous aggregations can be considered as primitive imitation of this multi-channelling mechanism.

\begin{table}
	\centering
	\caption{Convergence Variance for ER Curves}
	\begin{threeparttable}
	\centering
	\resizebox{.75\textwidth}{!}{
	\begin{tabular}{llllll}
		\toprule 
		Name         & Antivirus7k         & CYP2C9V12k          &  Mtb18k            & Malaria16k         &  Globalvirus76k \\
		\midrule
		GUNet        & $165.7\pm40.1$      & $215.4\pm39.7$      & $219.5\pm91.8$     & $342.4\pm229.6$    & $207.2\pm8.7$   \\
		GIN          & $225.2\pm15.0$      & $156.7\pm21.2$      & $181.1\pm34.6$     & $225.2\pm38.1$     & $157.0\pm3.9$ \\
		DeepChem     & $\ \ 62.0\pm15.0$   & $\ \ 36.1\pm4.7$    & $\ \ 28.9\pm5.4$   & $\ \ 16.1\pm1.9$   & $\ \ 15.6\pm1.6$   \\
		\cmidrule(r){1-6}
		HAG-Net cfg1 & $\ \ 66.6\pm14.4$   & $\ \ 65.6\pm4.2$    & $101.7\pm42.3$     & $\ \ 91.4\pm11.1$  & $\ \ 31.4\pm12.6$    \\
		HAG-Net cfg2 & $\ \ 76.1\pm41.1$   & $\ \ 62.9\pm24.3$   & $\ \ 45.2\pm3.6$   & $\ \ 42.5\pm3.7$   & $\ \ 24.5\pm7.3$   \\
		\bottomrule
	\end{tabular}}
	\label{tab:mstd}
	\end{threeparttable}
\end{table}

\section{Conclusion}

In this manuscript we improve GNN's performance from the perspective of enhancing information propagation from shallow to deep layers. As our first attempt in this direction, we propose to enhance information propagation by combining heterogeneous aggregation operators in GNN's neighborhood aggregation layers. By combining different aggregation operators, the information propagation loss can be mitigated, thus allowing more effective features for downstream task to propagate to deep layers. A new generic GNN layer formulation and upon this a new GNN variant referred as HAG-Net is proposed. We empirically validate the effectiveness of HAG-Net on a number of graph classification datasets. Our future work will investigate enhancing information propagation for GNN in the perspective of channelling as done in CNN.

\bibliographystyle{unsrtnat}
\bibliography{references.bib}  
\end{document}